\documentclass[11pt,letter]{article}

\usepackage[utf8]{inputenc}
\usepackage[T1]{fontenc}
\usepackage{amsmath,amssymb,amsfonts}
\usepackage{graphicx}
\usepackage[margin=1in]{geometry}
\usepackage{setspace}
\usepackage{hyperref}
\usepackage{cite}
\usepackage{siunitx}
\usepackage{lineno}

\hypersetup{
    colorlinks=true,
    linkcolor=blue,
    citecolor=blue,
    urlcolor=blue
}

\title{A Geometry-Informed Computer Vision Method for Detecting and Examining Overtaking Vehicles From A Bicycle}

\author{
    Gandhimathi Padmanaban\textsuperscript{1}, Rayane Moustafa\textsuperscript{1}, Fred Feng\textsuperscript{1,*}
}

\date{}

\begin{document}


\maketitle

\begin{center}
\small
\textsuperscript{1}Department of Industrial Manufacturing and Systems Engineering, University of Michigan-Dearborn, Michigan, USA, 48128\\
\textsuperscript{*}Corresponding author: fredfeng@umich.edu
\end{center}

\begin{abstract}
Instrumented bicycle studies have produced direct field evidence on vehicle
passing behavior, but the extraction of overtaking events from continuous
rear-facing video has remained dependent on manual, frame-by-frame annotation.
This bottleneck constrains sample sizes and limits the scope of naturalistic
cycling safety research. We present a geometry-informed computer vision pipeline
that automates overtaking event detection from a single bicycle-mounted camera
without requiring multi-sensor configurations or explicit camera calibration.
The system combines RT-DETR object detection with ByteTrack multi-object tracking
through a three-stage geometric validation module that enforces bearing angle
trend, apparent size growth, and spatial confirmation criteria derived from
perspective projection principles. Validated on 315 manually annotated real-world
overtaking events from urban roads in Ann Arbor, Michigan, the pipeline achieved
97.8\% recall with zero false positives. The system identified overtaking
intentions a mean of 2.44 seconds before vehicle passage, with 84.1\% of events
exceeding the 1.5-second human reaction time threshold, demonstrating feasibility
for active cyclist warning applications. Lateral passing distance measurements
from a subset of 96 events revealed that 33.3\% of passes occurred below the
commonly referenced 5-foot (152.4 cm) threshold, consistent with non-compliance
rates reported in prior field and self-reported studies. Additionally, a
preliminary calibration-free lateral distance estimation approach using bounding
box geometric features achieved mean absolute errors of approximately 13 to 14 cm
under leave-one-out cross-validation, sufficient to distinguish close passes from
standard passes for safety categorization purposes. By automating event isolation
from consumer-grade camera footage, the proposed system removes the primary
annotation bottleneck of instrumented bicycle research and provides a scalable
foundation for vehicle-bicycle interaction analysis across larger datasets and
diverse urban environments.

\vspace{3em}

\noindent \textbf{Keywords:} vehicle-bicycle interaction, geometry-informed
computer vision, overtaking detection, bicycle safety, lateral passing distance
\end{abstract}

\newpage

\section{Introduction}
\label{sec:intro}

Cyclist fatalities represent a persistent and growing challenge in transportation 
safety. The National Highway Traffic Safety Administration reported 1,166 bicycle 
fatalities and approximately 49,989 injuries in traffic crashes in the United 
States in 2023 alone~\cite{nhtsa_2023_cyclists}. Overtaking maneuvers, in which 
motorized vehicles pass cyclists from behind, account for a disproportionate share 
of severe injuries and fatalities among cyclists~\cite{nhtsa2022vulnerable, 
ntsb2019bicyclist, scarano2023systematic}. Understanding the spatial and temporal 
characteristics of these events is a prerequisite for designing effective 
interventions, whether through infrastructure modification, legal enforcement, or 
active warning systems. Yet the automated collection of reliable, event-level 
data from naturalistic cycling environments remains an unresolved methodological challenge.

\subsection*{Instrumented Bicycle Studies and the Annotation Bottleneck}

The instrumented bicycle paradigm has produced some of the most direct field 
evidence on passing behavior. Studies deploying ultrasonic sensors on bicycles 
have measured lateral passing distances under naturalistic conditions across 
multiple countries, consistently finding that a substantial proportion of passes 
occur below recommended safety thresholds~\cite{LOVE2012451, 
mackenzie2019passingdistances}. North American field measurements have reported 
mean lateral clearances in urban environments that are notably lower than European 
counterparts, a pattern attributed to narrower lane widths and higher traffic 
densities~\cite{chapman2012observations}. Despite their ecological validity, 
these studies share a fundamental limitation: identifying which segments of 
continuous footage correspond to actual overtaking events requires manual, 
frame-by-frame video review. This annotation bottleneck limits the volume of 
data that can realistically be processed, constrains sample sizes, and introduces 
inter-rater variability in event identification. An automated pipeline for 
isolating overtaking events from bicycle-mounted video would remove this 
constraint and enable data collection campaigns of substantially greater scope.

\subsection*{Computer Vision for Vehicle and Cyclist Safety}

Computer vision methods have advanced rapidly for vehicle and vulnerable road user 
detection. YOLO-family detectors~\cite{redmon2016yolo, jocher2022yolov5} and 
transformer-based architectures such as RT-DETR~\cite{zhaoDETR} have demonstrated 
high accuracy on standard benchmarks. Multi-object tracking algorithms, including 
ByteTrack~\cite{zhang2022bytetrack}, enable persistent vehicle trajectory 
estimation across video frames. These components have been applied to cyclist 
safety monitoring and vehicle conflict detection in various 
contexts~\cite{useche2024adas}. However, applying these tools to overtaking 
detection from bicycle-mounted cameras introduces challenges absent in 
infrastructure-based or vehicle-centric systems. The moving camera produces 
continuous ego-motion that complicates motion-based filtering. The relevant field 
of view is narrow and rapidly changing, with vehicles entering and exiting at 
high relative velocities. Critically, the large majority of vehicles visible in 
rear-facing footage are \emph{not} overtaking the bicycle: oncoming traffic, 
vehicles in parallel lanes, and vehicles that approach but decelerate or turn off 
all produce valid detection and tracking trajectories. Distinguishing completed 
overtaking events from this background requires semantic-level validation that 
appearance-based detectors alone do not provide.

Deep learning approaches address some of these challenges but introduce their own 
constraints. While YOLO-based systems achieve high accuracy in controlled 
environments~\cite{wang2019vehicle}, performance degrades significantly under 
environmental variations, camera movement, and non-standard viewing 
angles~\cite{zhang2020real}. Multi-sensor fusion systems attempt 
to address these limitations by combining cameras with lidar, radar, or 
ultrasound~\cite{yeong2021sensor}, but introduce temporal synchronization 
requirements, calibration dependencies, and hardware costs that limit deployment 
in low-resource research settings. Neither class of approach provides a principled 
method for distinguishing completed overtaking maneuvers from the broader 
population of vehicle trajectories captured in rear-facing bicycle footage.

\subsection*{Geometric Regularity of the Overtaking Scenario}

A critical and underexploited property of the overtaking scenario is its geometric 
regularity. As a vehicle approaches a cyclist from behind and moves alongside, its 
apparent position in the rear-facing image follows a trajectory governed by 
perspective geometry~\cite{hartley2004multiple, forsyth2012computer}. The 
vehicle's bearing angle from the camera reference point increases monotonically 
from approach to pass; its apparent size grows as it approaches and stabilizes as 
it moves ahead. These geometric signatures are consistent across camera models and 
mounting configurations, and they hold regardless of vehicle appearance, lighting 
conditions, or weather. Grounding event validation in these geometric invariants 
offers a route to detection robustness that does not depend on sensor-specific 
calibration or appearance-based learning.

\subsection*{Research Gap}

Despite a decade of instrumented bicycle research and rapid advances in computer 
vision, no validated automated pipeline exists for isolating overtaking events 
from bicycle-mounted rear-facing video. Existing detection systems either rely on 
appearance-based classifiers vulnerable to domain shift, require multi-sensor 
configurations with calibration and synchronization burdens, or lack the 
event-confirmation logic needed to distinguish completed overtaking maneuvers from 
the diverse vehicle trajectories present in naturalistic footage. The consequence 
is that the instrumented bicycle paradigm continues to depend on manual video 
review for event isolation, with directly constrains sample sizes and study scope. 
This gap motivates the present work.

\subsection*{Contributions}

This paper addresses these limitations through three contributions:

\begin{enumerate}

\item \textbf{A geometry-informed overtaking detection pipeline.} We present a 
three-stage system combining RT-DETR vehicle detection~\cite{zhaoDETR}, ByteTrack 
multi-object tracking~\cite{zhang2022bytetrack}, and a geometric validation module 
that enforces bearing angle trend, apparent size growth, and spatial confirmation 
criteria. The pipeline requires no camera calibration and is configurable for 
different mounting positions. Validated on 315 manually annotated real-world urban 
overtaking events from Ann Arbor, Michigan, the system achieves 97.8\% recall 
with zero false positives.

\item \textbf{Quantitative characterization of urban passing behavior.} Using 
automated event detections to isolate events for ultrasonic sensor analysis, we 
report lateral passing distance measurements for 96 events, finding that 33.3\% 
of measured passes fell below the 5-foot (152.4 cm) safety 
threshold~\cite{LOVE2012451, nhtsa2023countermeasures}. We further demonstrate 
that the system provides a mean advance detection time of 2.44 seconds prior to 
vehicle passage, with 84.1\% of events exceeding the 1.5-second human reaction 
time threshold~\cite{olson1986perception, Green01092000}.

\item \textbf{Preliminary calibration-free lateral distance estimation.} We 
demonstrate that bounding box geometric features produced by the detection 
pipeline, specifically the vertical position of the bounding box lower edge and 
apparent box area, carry sufficient information to estimate lateral passing 
distance without explicit camera calibration. Under leave-one-out cross-validation 
on 96 events with ultrasonic ground truth, mean absolute errors of approximately 
13 to 14 cm are achieved, sufficient to distinguish close passes from standard 
passes for safety categorization purposes.

\end{enumerate}

The remainder of this paper is organized as follows. Section~\ref{sec:methods} 
describes the data collection setup, system architecture, and geometric validation 
logic. Section~\ref{sec:results} presents detection performance, advance warning 
timing, passing behavior, and distance estimation results. 
Section~\ref{sec:discussion} interprets findings in relation to prior work, 
characterizes limitations, and identifies future directions.

\section{Method}
\label{sec:methods}

This section describes the geometry-informed computer vision system we developed to detect and analyze vehicle overtaking events from bicycle-mounted camera footage. The system processes 2D video sequences through three stages: vehicle detection, multi-object tracking, and geometric validation. We designed this pipeline to operate on single-camera data while maintaining robust performance across varying urban traffic conditions.

\subsection{Data Collection and Experimental Setup}
\begin{figure}[]
    \centering
    \includegraphics[width=0.75\linewidth]{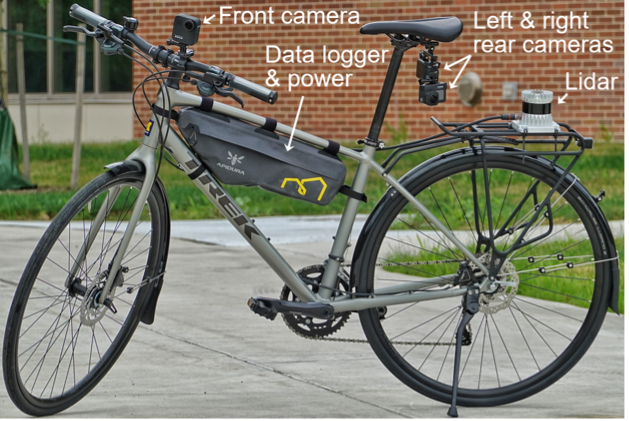}
    \caption{The research bicycle equipped with multiple cameras and sensors.}
    \label{fig:bike}
\end{figure}

We collected data using an instrumented research bicycle, shown in Figure \ref{fig:bike}, equipped with multiple cameras and sensors. The bicycle platform carried four GoPro cameras recording at 1920×1080 resolution positioned to provide coverage around the bicycle, along with a LiDAR and a C3FT device for distance measurements.

For this work, we only analyzed footage from the rear-left facing GoPro camera. This position captures vehicles approaching from behind, the relevant view angle for detecting overtaking maneuvers, while minimizing occlusion from the rider and bicycle frame.

We collected two trips of video on urban roads in Ann Arbor, Michigan. The routes consisted of dedicated bicycle lanes on roadways with two travel lanes in each direction, capturing data across different traffic densities and conditions. The dataset contains 315 manually verified vehicle overtaking events, defined as motorized vehicles (cars, trucks, buses) that approached from behind, passed alongside the bicycle, and moved ahead of the bicycle.

Ground truth was established through frame-by-frame manual annotation by the research team. For a subset of 96 events, we also obtained lateral passing distance measurements using a C3FT ultrasonic sensor to validate the spatial accuracy of our geometric calculations.

\subsection{System Architecture}

\begin{figure}[]
    \centering    \includegraphics[width=\columnwidth]{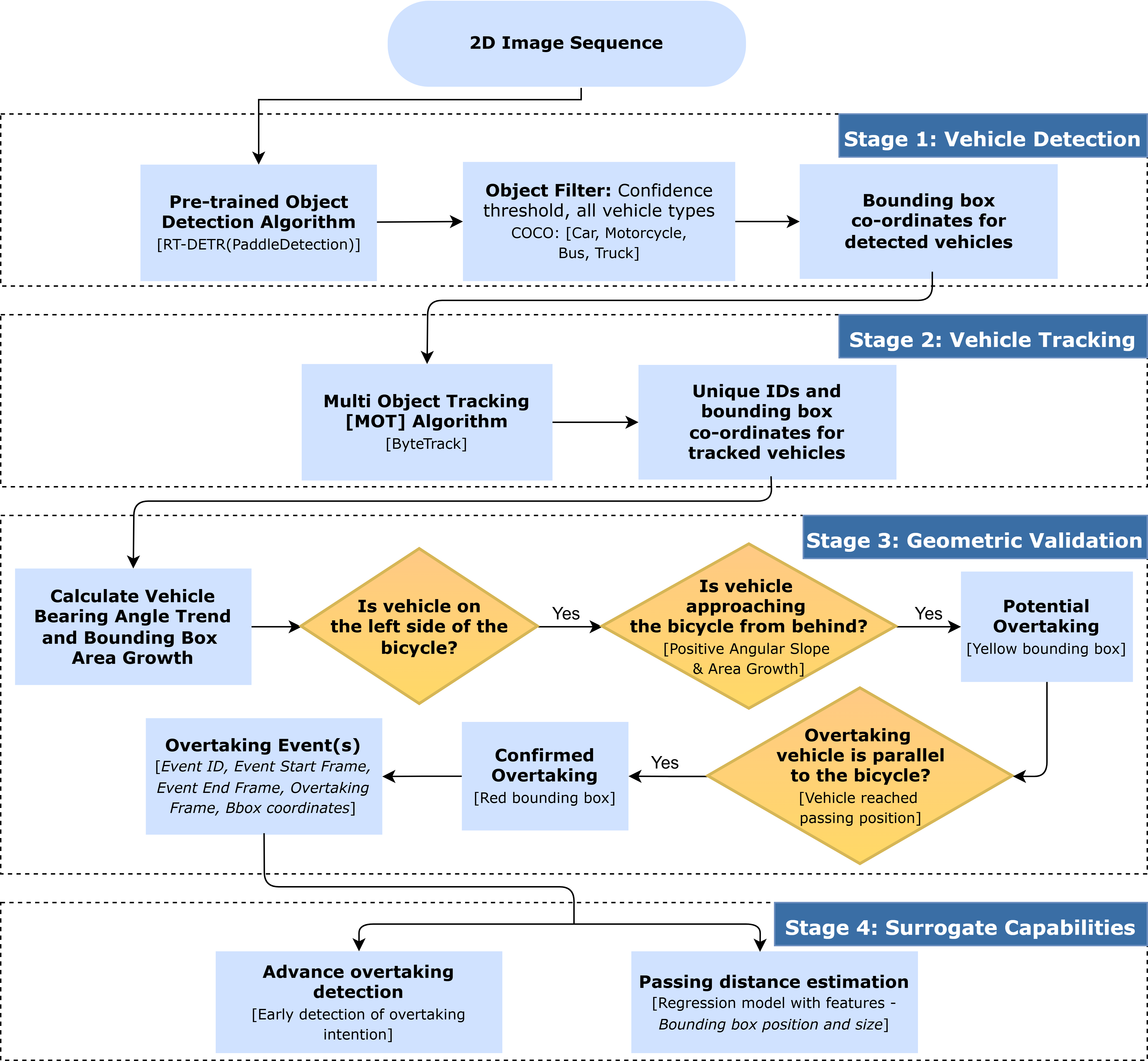}
    \caption{System architecture showing the three-stage geometry-informed overtaking detection
pipeline: vehicle detection, vehicle tracking, and geometric validation.}
    \label{fig:sys_architecture}
\end{figure}

The detection pipeline consists of three sequential stages. First, an object detection model identifies vehicles in each frame. Second, a tracking algorithm links detections across frames into continuous trajectories. Third, a geometric validation module analyzes each trajectory to determine whether it represents a completed overtaking event. Figure \ref{fig:sys_architecture} illustrates this processing flow.

We implemented the system to run on standard computing hardware using ONNXRuntime for model inference. The modular design allows the detector, tracker, and validation components to be updated or replaced independently. The geometric features extracted during this validation process (e.g., bounding box size and position) serve as inputs for downstream safety analysis, such as the preliminary distance estimation demonstrated in the Results section.

\subsubsection{Vehicle Detection}
We use RT-DETR (Real-Time Detection Transformer) \cite{zhaoDETR} as the object detection model. RT-DETR is a transformer-based detector that processes images through an efficient hybrid encoder and uses IoU-aware query selection for object localization. Compared to convolutional detectors like YOLO \cite{redmon2016yolo,jocher2022yolov5}, the transformer architecture provides better handling of scale variations and partial occlusions which is a common challenge in traffic video from moving bicycles.

We deployed the RT-DETR-l model from the PaddleDetection framework, exported to ONNX format for cross-platform compatibility. The detector identifies four vehicle classes from the COCO dataset \cite{LinCOCO}: car, motorcycle, bus, and truck.

A practical issue we encountered was duplicate detections, where the same vehicle sometimes receives multiple bounding boxes with different class labels. These duplicates can cause tracking errors by creating multiple track IDs for a single vehicle. We implemented a pre-tracking suppression step that removes near-duplicate boxes based on high intersection-over-union overlap, retaining only the highest-confidence detection for each vehicle.

\subsubsection{Multi-Object Tracking}
Linking detections into continuous trajectories requires handling occlusions, camera motion, and the high relative velocities typical of vehicle-bicycle interactions. We use ByteTrack \cite{zhang2022bytetrack} for this tracking task. ByteTrack maintains track identity by matching detections across frames using position and appearance similarity, while using low-confidence detections to recover tracks that become temporarily occluded.

We adjusted ByteTrack's parameters from their default settings to accommodate the vehicle overtaking scenario. Specifically, we raised the track activation threshold to initialize tracks only from high-confidence detections, reducing false starts from background objects. We also reduced the matching threshold to allow looser spatial matching during rapid position changes, and limited the lost track buffer duration to prevent ID carryover between different vehicles.

We also observed that ByteTrack occasionally assigns multiple IDs to the same vehicle during brief occlusions or when detection quality fluctuates. To address this, we implemented a same-frame duplicate filter that checks if any two track IDs have overlapping bounding boxes and similar geometric properties. When duplicates are detected, we retain the track with higher detection confidence.

\subsubsection{Geometric Validation of Overtaking Events}
In urban environments with four-lane roadways, a rear-facing bicycle camera captures numerous vehicles that do not represent overtaking events. Vehicles traveling in oncoming lanes, sitting stationary at traffic signals, parked along the roadside, or approaching from behind but turning off before passing all produce valid detections and tracking trajectories. The continuous motion of the bicycle further complicates this picture, as vehicles may enter and exit the frame at various positions depending on the bicycle's lane position and the surrounding traffic flow. The geometric validation stage filters these diverse scenarios to identify only those trajectories that correspond to completed overtaking maneuvers (vehicles that approached from behind, moved alongside the bicycle, and proceeded ahead).

\textbf{Reference Frame and Angle Calculation:} We compute the bearing angle of each vehicle relative to a fixed reference point in the image. For the rear-left camera position, the reference point is set at the bottom-left corner of the frame \((x_{ref} = 0, y_{ref} = H)\), where \(H\) is the frame height. This anchors the calculation to the road surface rather than variable vehicle features. The reference point location is configurable based on the camera's lateral mounting position (e.g., bottom-center for a rear-center camera), while remaining anchored to the frame's bottom edge to maintain geometric consistency.

The bearing angle $\theta$ is calculated as:

\[ \theta = \operatorname{atan2}\!\left(x_{vehicle} - x_{ref},\; y_{ref} - y_{vehicle}\right) \times \frac{180}{\pi} \]

where ($x_{vehicle}, y_{vehicle}$) is the centroid of the vehicle's bounding box. For the rear-left camera position, overtaking vehicles approach from the left side of the frame and move rightward during the pass, producing bearing angles that increase from 0° to 90°.

\textbf{Probationary Validation Protocol:} A tracked vehicle enters a probationary evaluation phase once it appears in the valid angle range. During probation, the system maintains a sliding window history of the vehicle's bearing angle and bounding box area. Three conditions must be satisfied simultaneously for the track to advance to potential overtaking status:
\begin{enumerate}
    \item \emph{Positive angular trend:} Overtaking vehicles move from left to right across the frame, producing increasing bearing angles. We fit a linear regression to the angle history and require the regression slope to exceed a minimum threshold. Vehicles exhibiting a consistently decreasing slope—indicative of oncoming traffic or a ByteTrack ID switch to a different vehicle—are removed from consideration.
    \item \emph{Apparent size growth:} As a vehicle approaches from behind, perspective projection causes its apparent size to increase. We calculate the rate of change in bounding box area over the history window. Only tracks whose area growth rate exceeds a minimum threshold continue evaluation, filtering out static background objects and vehicles that are not approaching.
    \item \emph{Temporal consistency:} To be promoted from probation to potential overtaking, a track must maintain positive angular trend and size growth for a minimum number of consecutive frames. This duration requirement eliminates transient detections and tracking noise.
\end{enumerate}

Tracks that pass probation are marked as potential overtaking events. We continue monitoring their geometry frame-by-frame. If the angular trend reverses or size growth becomes negative, we remove the track from consideration, assuming the ID has been reassigned to a different vehicle. Only spatially confirmed events (described below) are recorded in the final output.

\textbf{Spatial Confirmation:} A potential overtaking becomes a confirmed event when the right edge of the vehicle's bounding box crosses a spatial threshold. This threshold is a configurable parameter based on camera position. For the rear-left setup used in this study, it is set at 55\% of the frame width. This ensures the vehicle has moved sufficiently across the frame to constitute a completed pass, filtering out vehicles that approach but change lanes, slow down, or otherwise fail to complete the maneuver.

When a track crosses the confirmation threshold, we record the event with its associated metadata: track ID, first and last frame, vehicle class, confirmation frame, bearing angle, and bounding box coordinates at the moment of confirmation.

This validation approach relies on perspective geometry principles that remain stable across different camera types and mounting positions, requiring only adjustment of the confirmation threshold, angle range and reference point location for different setups and traffic configurations.

\section{Results}
\label{sec:results}

\subsection{System Performance}
The system successfully detected and confirmed 308 of 315 ground truth overtaking events, yielding a recall of 97.8\%. Notably, the system produced zero false positives across both collection trips, indicating that the three-stage geometric validation effectively filters non-overtaking vehicle trajectories without generating spurious detections. Performance remained consistent across both collection trips, with Trip 1 achieving 99.3\% recall (138 of 139 events) and Trip 2 achieving 96.6\% recall (170 of 176 events). The modest difference between trips reflects variations in traffic density and lighting conditions encountered during different times of day.

The seven missed detections fell into two categories. Four events (57\%) were caused by occlusion from other passing vehicles. In these cases, a vehicle in the nearer lane temporarily blocked the camera's view of a second vehicle overtaking in the farther lane, preventing the occluded vehicle from accumulating sufficient tracking history to pass the geometric validation criteria. The remaining three events (43\%) resulted from inconsistent RT-DETR detections where the vehicle was either not detected during the critical approach phase or detected only sporadically after it had already passed the spatial confirmation threshold. These intermittent detections prevented the formation of continuous tracks required for the probationary validation process.

The detected events spanned the expected vehicle types: 259 cars (84.1\%), 46 trucks (14.9\%), and 3 buses (1.0\%). No motorcycles were present in this dataset, though the system is configured to detect them.

\subsection{Early Warning Performance}
For safety intervention applications, the time between initial detection and the moment a vehicle draws alongside the bicycle determines whether a warning system can provide actionable alerts. We measured advance warning time as the interval between the frame when a track entered the probationary validation phase (exhibiting geometric patterns consistent with overtaking behavior) and the frame when it crossed the spatial confirmation threshold.

\begin{figure}[]
    \centering
    \includegraphics[width=\textwidth]{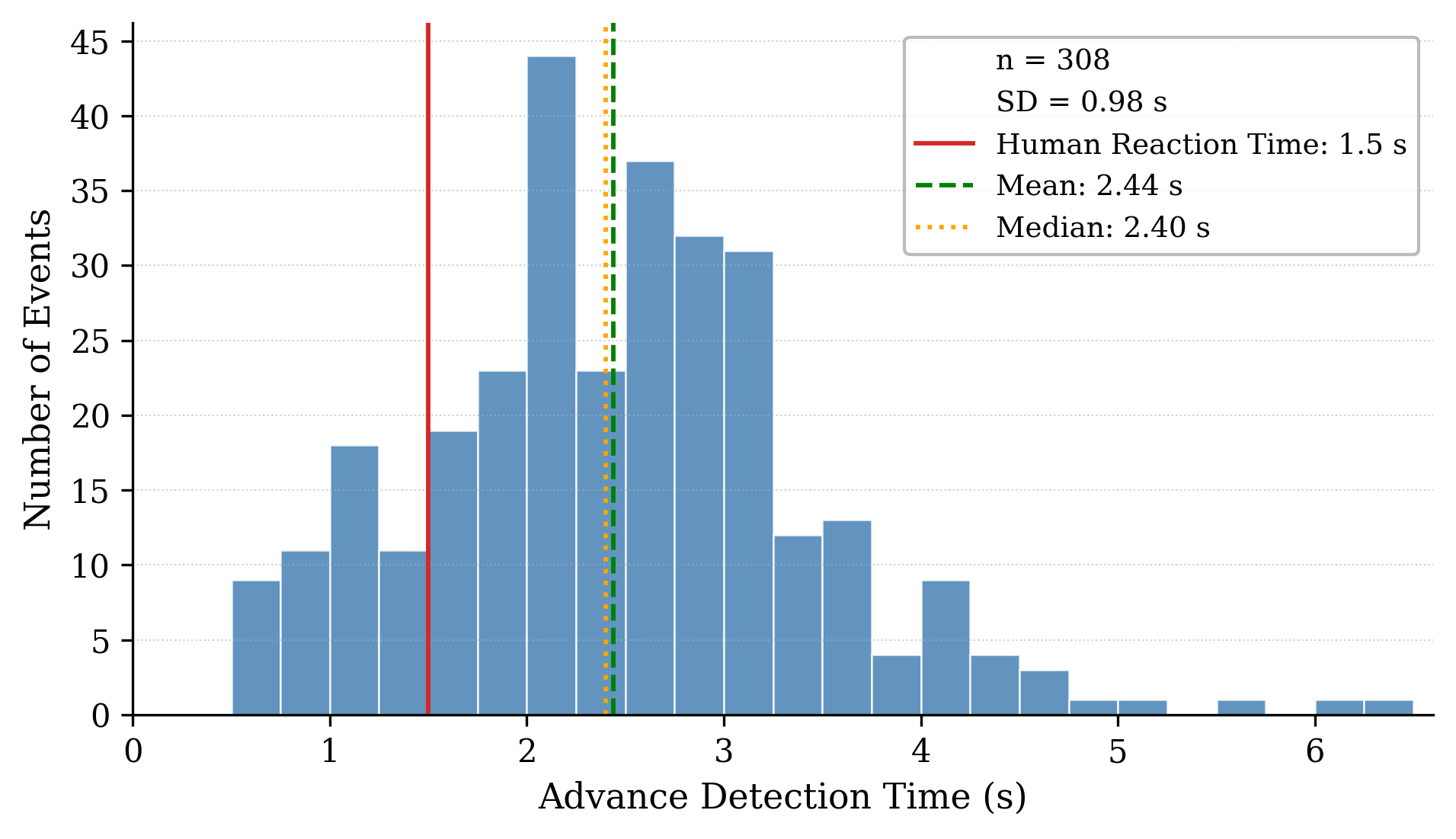}
    \caption{Distribution of advance detection times of overtaking intention.}
    \label{fig:advance_detection}
\end{figure}

Across the 308 detected events, the system identified overtaking intentions an average of 2.44 seconds (median: 2.40 seconds, SD: 0.98 seconds) before vehicles drew alongside the bicycle. The distribution ranged from 0.50 to 6.30 seconds, with the majority of events (84.1\%) meeting or exceeding the 1.5-second human reaction time threshold commonly cited in driver response studies \cite{olson1986perception, Green01092000}. Figure \ref{fig:advance_detection} illustrates the distribution of advance detection times across all detected events.

This temporal buffer provides sufficient lead time for active safety interventions. A cyclist receiving a 2.4-second advance notification has time to adjust lane position, check over their shoulder, or prepare for turbulence from a large vehicle. The consistency of this metric across diverse traffic scenarios suggests that the geometric progression patterns emerge predictably during the approach phase, well before vehicles reach critical proximity.

\subsection{Observed Passing Behavior}
The C3FT ultrasonic sensor provided lateral passing distance measurements for 96 of the 315 overtaking events. This subset represents the events where vehicles passed within the sensor's effective ranging distance. These measurements enabled quantitative analysis of real-world passing behavior in urban cycling environments.

\begin{figure}[]
    \centering
    \includegraphics[width=\textwidth]{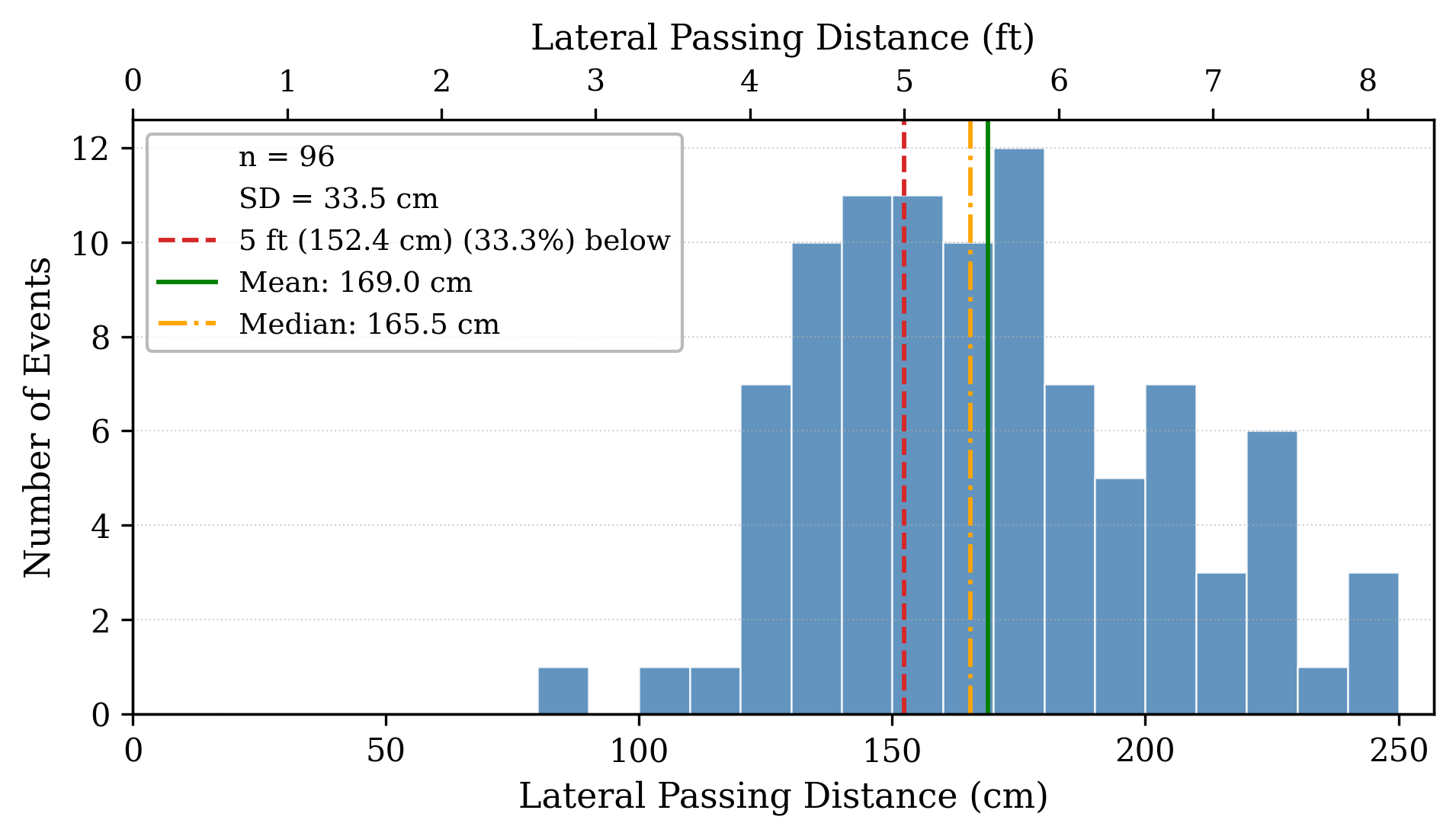}
    \caption{Distribution of lateral passing distances from C3FT device.}
    \label{fig:passing_distance}
\end{figure}

The measured passing distances showed substantial variability  (Figure \ref{fig:passing_distance}). The mean lateral clearance was 169.0 cm (5.5 feet, SD: 33.5 cm), with distances ranging from 88.0 cm to 247.0 cm. The distribution revealed that one-third of measured passes (32 of 96 events, 33.3\%) occurred at distances below 152.4 cm (5 feet), a threshold commonly referenced in traffic safety guidelines and legal minimum passing distance requirements in multiple jurisdictions \cite{LOVE2012451, nhtsa2023countermeasures}.

At the lower extreme, one event recorded a passing distance of 88.0 cm (2.9 feet), representing a clearance significantly below recommended safety buffers. The median passing distance of 165.5 cm (5.4 feet) fell only marginally above the 5-foot threshold, indicating that typical passing behavior in these urban environments operates near commonly recommended safety margins.

\subsection{Vision-Based Distance Estimation}
The bounding box geometry produced at detection time carries information about lateral passing distance. Perspective projection dictates that a vehicle passing at closer range occupies a larger image footprint with its lower edge displaced toward the bottom of the frame. These projective relationships allow lateral distance to be estimated from a single camera frame without explicit calibration: the inverse bounding box height, the normalized lower-edge position, and the box aspect ratio serve as predictors, with the normalized lower-edge position providing the dominant signal consistent with the geometric interpretation on flat roadways.

\begin{figure}[]
    \centering
    \includegraphics[width=\textwidth]{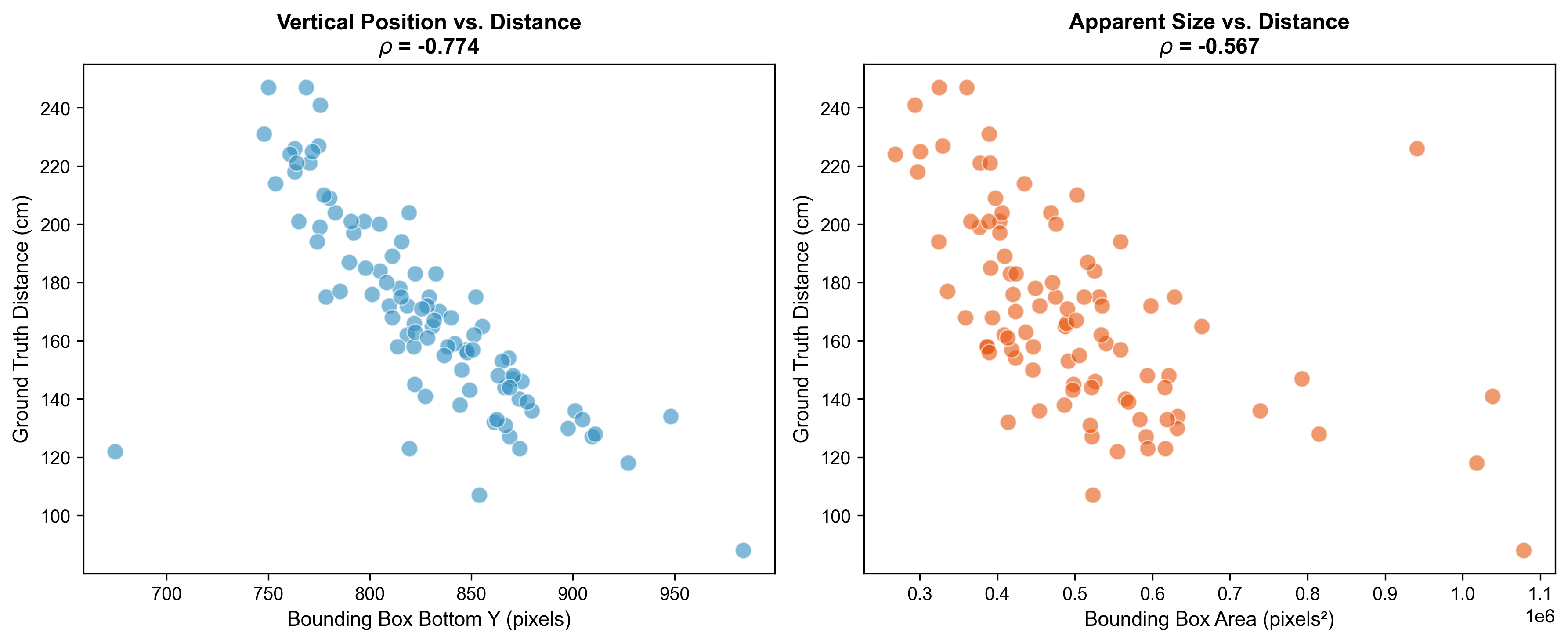}
    \caption{Spearman correlation between bounding box geometric features and measured lateral passing distance for 96 events with C3FT ground truth. Vertical position of the bounding box bottom edge ($\rho = -0.774$, left) is the dominant predictor, consistent with perspective geometry on flat roadways. Bounding box area ($\rho = -0.567$, right) provides a secondary apparent-size cue whose signal strength varies with vehicle class.}
    \label{fig:distance_correlations}
\end{figure}

Analysis of the 96 events with available ground truth C3FT measurements confirms these geometric expectations quantitatively. The vertical position of the bounding box's bottom edge shows the strongest correlation with measured passing distance ($\rho = -0.774$), while bounding box area provides a secondary signal ($\rho = -0.567$). Other candidate features, including horizontal position and passing angle, exhibited weak correlations ($|\rho| < 0.06$) and were excluded. This ordering is reflected in the fitted model coefficients: vertical position accounts for 77\% of the predictive weight, with bounding box area contributing the remaining 23\%. Figure~\ref{fig:distance_correlations} illustrates these feature--distance relationships across the measurement subset.

\begin{figure}[]
    \centering
    \includegraphics[width=0.65\linewidth]{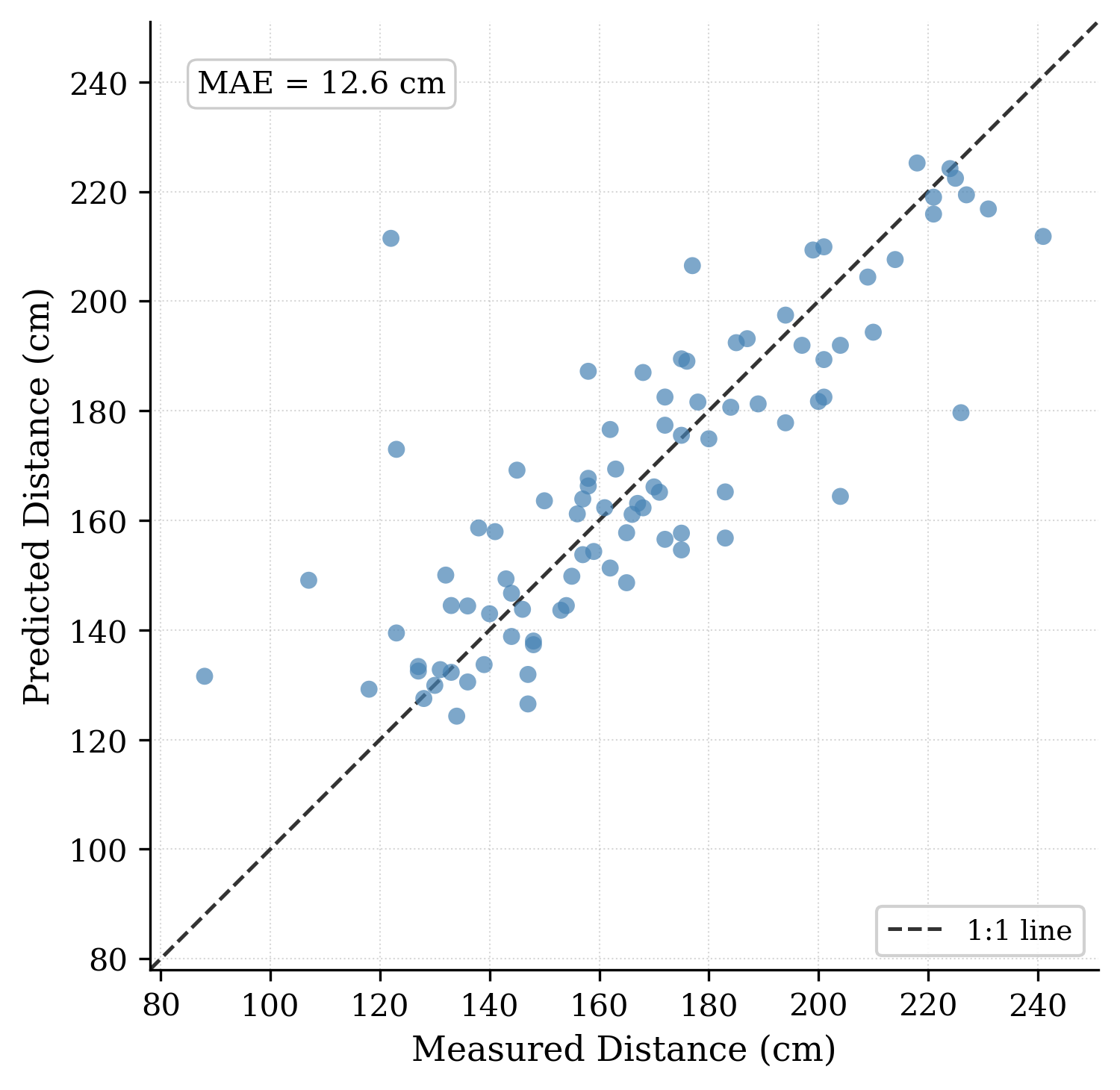}
    \caption{Predicted versus measured lateral passing distance under leave-one-out cross-validation. The dashed line is the 1:1 reference; the dotted lines mark the 152.4 cm (5 ft) threshold. Points below and to the right of the threshold intersection correspond to close passes that the model correctly identifies as such.}
    \label{fig:distance_scatter}
\end{figure}

Using the detected events with available C3FT ground truth measurements (lateral range: 0.88--2.41 m), preliminary evaluation under leave-one-out cross-validation yields a mean absolute error of approximately 13--14 cm regardless of regression approach, sufficient to distinguish close passes (under 1.5 m) from standard passes (over 2 m) for safety categorization (Figure \ref{fig:distance_scatter}). A detailed evaluation, including comparison of regression approaches, uncertainty quantification, and generalization testing, is the subject of ongoing work.

More broadly, the automated detection pipeline removes the manual review step that  has historically constrained the scope of instrumented bicycle studies. The detection system identifies when and where events occur, enabling efficient extraction of bounding box features and corresponding sensor measurements without frame-by-frame video inspection. The system therefore supports both sensor-augmented and sensor-free behavioral analysis across larger future datasets.

\section{Discussion}
\label{sec:discussion}

\subsection{Detection Performance}

The geometry-informed pipeline achieved 97.8\% recall with zero false positives 
across 315 manually verified overtaking events. The absence of false positives is 
particularly notable: it reflects the specificity of the three-stage geometric 
validation, which requires simultaneous satisfaction of angle trend, bounding box 
growth, and spatial confirmation criteria before an event is recorded. This is a 
meaningful distinction from deep learning classifiers trained on appearance 
features, which frequently produce false alarms under domain 
shift~\cite{zhang2020real,wang2019vehicle}. The low false positive rate is 
practically important for safety applications where spurious alerts erode user 
trust and system adoption.

The seven missed events highlight two distinct failure modes that are 
architecturally separate from the geometric validation itself. The first failure mode, occlusion by a nearer-lane vehicle blocking the camera's 
view of a farther-lane overtaker, is an inherent limitation of single-viewpoint 
systems and has been documented in cyclist overtaking studies using fixed camera 
configurations~\cite{TOULOUSE20253361}. The second failure mode, intermittent 
RT-DETR detections preventing sufficient tracking history, reflects a 
detection-layer constraint rather than a validation-layer failure. Both failure 
modes are theoretically addressable without modifying the geometric logic: 
multi-angle camera coverage would mitigate occlusion, while ensemble detection 
or higher frame rates could improve track continuity.

Compared to panoramic and multi-camera approaches for cycling 
monitoring~\cite{guo2026multipleobjectdetectiontracking}, the single rear-left configuration used 
here accepts a narrower field of view in exchange for lower hardware cost and 
simpler deployment. This trade-off is appropriate for research contexts 
prioritizing scalability and reproducibility over full-scene coverage.

\subsection{Passing Distance Findings and Policy Relevance}

Finding that one-third of measured passes occurred below the 5-foot (152.4 cm) 
threshold is consistent with prior field studies. Chapman and 
Noyce~\cite{chapman2012observations} and the NHTSA countermeasures 
report~\cite{nhtsa2023countermeasures} document that drivers routinely 
underestimate safe lateral clearance, particularly on roads with marked bicycle 
lanes where the lane marking may create a false sense of separation. The mean lateral clearance of 169.0 cm in this dataset is consistent with values 
reported in prior instrumented bicycle studies, which typically range from 
approximately 120 to 200 cm depending on road type, jurisdiction, and traffic 
conditions~\cite{LOVE2012451, mackenzie2019passingdistances, llorca2017motorist}. 
The variability across these studies underscores that passing behavior is sensitive 
to local infrastructure, legal context, and traffic composition, and that 
site-specific measurement remains necessary for meaningful safety assessment.

The distribution tail is particularly concerning: a minimum recorded pass of 88.0 cm represents a clearance that would likely be insufficient to avoid a collision in the event of any lateral deviation by either the cyclist or the driver. One-third of passes fell below the 5-foot guideline commonly referenced 
in transportation safety literature and adopted as a legal minimum in multiple US jurisdictions~\cite{LOVE2012451,nhtsa2023countermeasures}. This rate of non-compliance is consistent with field-measured non-compliance rates 
of approximately 15\% in Australian urban environments~\cite{DEBNATH2018137} 
and self-reported non-compliance rates of approximately 35\% in similar 
jurisdictions~\cite{HAWORTH2018183}, collectively supporting prior 
findings that legal minimums alone, without enforcement or physical infrastructure 
modifications, do not reliably achieve safe passing 
margins~\cite{nhtsa2023countermeasures, chapman2012observations,SINCLAIR2026100102}. The ability to collect quantitative passing behavior evidence from a single 
consumer-grade camera, without manual video review, is a direct practical 
contribution of the proposed system.

\subsection{Advance Warning Timing and Safety Intervention Feasibility}

The 2.44-second mean advance detection time provides an actionable temporal 
buffer for cyclist warning systems. Olson and Sivak's work on driver 
perception-reaction time documents a minimum of approximately 1.5 seconds under 
alerted conditions~\cite{olson1986perception,Green01092000}, and recent studies 
of cyclist-specific perception-reaction time report an 85th percentile of 
approximately 0.84 seconds for unexpected 
hazards~\cite{martin2025cyclistreaction}. The observation that 84.1\% of detected 
events exceeded the 1.5-second threshold suggests that the geometric progression 
of overtaking, the steady angular migration and bounding box expansion, provides 
a natural predictive signal that precedes the physical passing moment by a margin 
sufficient for intervention in the majority of cases.

This compares favorably to in-vehicle forward collision warning systems reviewed 
by Useche et al.~\cite{useche2024adas}, which typically provide 1 to 2 
seconds of warning for imminent cyclist conflicts. A rear-mounted bicycle warning 
system capable of consistent 2.44-second advance detection operates in a 
comparable or superior temporal window, with the advantage of alerting the 
cyclist rather than the overtaking driver. However, events with advance warning 
under one second, associated with fast-approaching vehicles and occasional 
tracking instability during the approach phase, represent edge cases where 
geometric progression alone is insufficient and additional sensing or logic would 
be required.

\subsection{Vision-Based Distance Estimation}

The preliminary distance estimation results confirm that geometric bounding box 
features carry meaningful information about lateral passing distance, consistent 
with well-established perspective projection 
theory~\cite{hartley2004multiple,forsyth2012computer}. The dominance of the 
bounding box bottom-edge position (77\% of model weight) over apparent area (23\%) 
is expected on flat roadways, where the ground contact point varies predictably 
with lateral distance. This ordering would likely shift on roads with significant 
cross-slope or where vehicles occupy different lateral offsets, an important 
caveat for generalization beyond the tested environment.

The mean absolute error of approximately 13 to 14 cm under leave-one-out 
cross-validation is competitive with monocular ranging methods that rely on 
explicit camera calibration. Bounding-box-based ranging without calibration has 
been reported at errors of approximately 10 to 16\% of range for forward-facing 
highway applications~\cite{usmankhujaev2023distance}; our lateral results, spanning a 
range of \SIrange{0.88}{2.41}{\metre}, fall within a comparable relative error 
envelope. The key distinction is that the present approach requires no calibration 
and derives all predictors from the same detection bounding boxes used for event 
identification, which simplifies deployment compared to systems that require 
separate calibration procedures or dedicated ranging hardware.

This estimation capability is explicitly presented as preliminary. The sample of 
96 events with C3FT ground truth is modest, the training range does not extend to 
passing distances beyond approximately 2.4 m, and the linear regression model 
makes strong assumptions about the distance-geometry relationship. A full evaluation, including analysis of vehicle class effects, 
out-of-distribution generalization, and uncertainty quantification, is 
the subject of ongoing work.

\subsection{Limitations}

Several limitations constrain the generalizability of these findings. First, all 
data were collected on a single urban route in Ann Arbor, Michigan, with two 
travel lanes in each direction and dedicated bicycle lanes. Performance on routes 
with different lane configurations, roadway geometries, or traffic compositions 
has not been evaluated. Second, the dataset was collected under daylight 
conditions with standard weather; the system's behavior in low-light, adverse 
weather, or high-density traffic scenarios remains untested. While the geometric 
validation logic is theoretically invariant to illumination, the upstream RT-DETR 
detector is not, and detection failures under poor visibility would propagate 
directly to the validation stage.

Third, the confirmation threshold and reference point location are tuned for the 
rear-left camera position used in this study. Applying the system to other bicycle 
configurations or non-standard camera mounts requires parameter re-adjustment, 
and the sensitivity of performance to those adjustments has not been 
characterized. Fourth, the distance estimation model was validated only against 
the C3FT ultrasonic sensor, which has its own measurement limitations including 
sensitivity to vehicle surface geometry and angle of incidence. An independent 
validation against a calibrated reference instrument would strengthen the 
distance estimation claims. Finally, the dataset does not include night-time 
riding, high-speed roads, or multi-lane highway environments, where different 
geometric relationships and faster approach velocities may challenge both the 
probationary validation timing and the advance warning margins documented here.

\subsection{Implications for Cycling Safety Research and Practice}

The system directly addresses a fundamental scalability bottleneck in naturalistic 
cycling safety research. Instrumented bicycle studies with ultrasonic distance 
sensors have been conducted for over a 
decade~\cite{LOVE2012451,mackenzie2019passingdistances}, but the extraction of passing 
events from continuous video has remained largely manual, limiting the scope of 
behavioral analysis achievable within practical annotation budgets. The proposed 
pipeline automates event detection from a single consumer-grade camera, removing 
the manual review step that has historically constrained sample sizes and enabling 
future deployment across multiple riders, routes, and data collection campaigns 
without proportionally increasing annotation effort.

The passing distance findings also carry direct policy relevance. The documented 
rate of non-compliance with the 5-foot passing 
guideline~\cite{LOVE2012451,nhtsa2023countermeasures} on a bicycle-lane corridor 
provides quantitative evidence that bicycle lane markings alone do not reliably 
produce compliant driver behavior. Such evidence can inform enforcement 
prioritization, infrastructure design decisions, and public awareness campaigns. 
The ability to generate this evidence from low-cost consumer cameras lowers the 
barrier for jurisdictions to conduct their own compliance monitoring studies, 
which has been identified as a critical gap in cycling policy 
evaluation~\cite{nhtsa2023countermeasures}.

\subsection{Future Research Directions}

Several extensions follow directly from this work. Validation across diverse 
routes, lighting conditions, and camera configurations is the most pressing need; 
deploying the system across multiple riders and cities would characterize both 
generalizability and the variability of urban passing behavior. Expanding the 
distance estimation dataset, particularly with passes beyond 2.4 m and with 
independent calibrated ground truth, would enable a rigorous evaluation of the 
preliminary results reported here, including quantification of vehicle class 
effects and cross-route generalization.

Integration of rear- and side-facing cameras could address the occlusion failure 
mode and enable simultaneous characterization of the cyclist's lateral lane 
position, which influences driver behavior and is currently unmeasured. The 
geometric framework is also extensible to other vulnerable road user contexts: 
pedestrian-vehicle lateral proximity at crossings and dooring risk for cyclists 
near parked vehicles share structural similarities with the overtaking geometry 
analyzed here. Finally, real-time deployment as a cyclist warning device, rather 
than a post-hoc analysis tool, would require evaluation of system latency and 
energy consumption on edge computing hardware, representing a practical bridge 
between the research contributions reported here and operational safety 
applications.

\section{Conclusions}
\label{sec:conclusions}

This paper presented a geometry-informed computer vision pipeline for automated
detection and characterization of vehicle overtaking events from bicycle-mounted
camera footage. The system integrates RT-DETR object detection, ByteTrack
multi-object tracking, and a three-stage geometric validation module that enforces
bearing angle trend, apparent size growth, and spatial confirmation criteria
derived from perspective projection principles. Validated on 315 manually
annotated real-world overtaking events collected in Ann Arbor, Michigan, the
pipeline achieved 97.8\% recall with zero false positives, confirming that
geometric invariants provide a reliable and specific basis for event identification
without dependence on appearance-based classifiers or multi-sensor configurations.

The system produced a mean advance detection time of 2.44 seconds prior to vehicle
passage, with 84.1\% of events exceeding the 1.5-second human reaction time
threshold. This temporal margin is sufficient for active safety interventions and
compares favorably with in-vehicle forward collision warning systems, while
offering the additional advantage of alerting the cyclist rather than the
overtaking driver. Lateral passing distance measurements from a subset of 96
events revealed that one-third of passes occurred below the commonly referenced
5-foot (152.4 cm) threshold, a rate of non-compliance consistent with prior
instrumented field studies and self-reported surveys conducted in comparable
jurisdictions. These findings provide quantitative, site-specific evidence that
bicycle lane markings alone do not reliably produce compliant driver behavior.

The preliminary calibration-free distance estimation component demonstrated that
bounding box geometric features, specifically the vertical position of the
bounding box bottom edge and apparent box area, carry sufficient information to
estimate lateral passing distance with a mean absolute error of approximately
13 to 14 cm under leave-one-out cross-validation. This result is competitive with
monocular ranging methods that require explicit camera calibration, and it derives
entirely from the same detection output used for event identification, requiring
no additional hardware or calibration procedures.

Collectively, these contributions address the annotation bottleneck that has
historically constrained the scope of instrumented bicycle studies. The pipeline
automates event isolation from a single consumer-grade camera, enabling future
data collection campaigns across multiple riders, routes, and traffic environments
without proportional increases in manual annotation effort. The modular
architecture allows individual components to be updated independently, and the
geometric validation logic requires only parameter adjustment for deployment on
different camera configurations. Validation across diverse routes, lighting
conditions, and camera mounts represents the most immediate direction for future
work, along with expansion of the distance estimation dataset and integration of
additional camera viewpoints to address the occlusion failure mode identified
here.

\section{Acknowledgments}
\begin{itemize}
    \item This work is supported by the National Science Foundation under award number 2142757.
    \item Grammarly was occasionally used for grammar and spelling check.
\end{itemize}

\bibliographystyle{plain} 
\bibliography{ref} 

\end{document}